\documentclass[10pt, a4paper]{article}

\usepackage{lrec-coling2024} % this is the new style
\usepackage{graphicx}
\usepackage{tabularx}
\usepackage{soul}
\usepackage{booktabs}
\usepackage{multirow}
\usepackage[T1]{fontenc}
\usepackage{CJK}
\usepackage[utf8]{inputenc}
\usepackage{makecell}
\usepackage{utfsym}
\usepackage{xcolor}

\title{FineMath: A Fine-Grained Mathematical Evaluation Benchmark for Chinese Large Language Models\\}

\name{Yan Liu\textsuperscript{\rm{1}}, Renren Jin\textsuperscript{\rm{1}}, Ling Shi\textsuperscript{\rm{2}}, Zheng Yao\textsuperscript{\rm{3}}, Deyi Xiong\textsuperscript{\rm{1}}\sthanks{~Corresponding author.}} 

\address{\textsuperscript{1}Tianjin University, Tianjin, China,\\
        \textsuperscript{2}China University of Geosciences, Wuhan, China,\\
        \textsuperscript{3}The University of Queensland, QLD, Australia,\\
         \{yan\_liu,rrjin,dyxiong\}@tju.edu.cn\\
         \{lingshi0265\}@gmail.com\\
         \{zheng.yao1\}@uq.net.au\\}

\abstract{
To thoroughly assess the mathematical reasoning abilities of Large Language Models (LLMs), we need to carefully curate evaluation datasets covering diverse mathematical concepts and mathematical problems at different difficulty levels. In pursuit of this objective, we propose FineMath in this paper, a fine-grained mathematical evaluation benchmark dataset for assessing Chinese LLMs. FineMath is created to cover the major key mathematical concepts taught in elementary school math, which are further divided into 17 categories of math word problems, enabling in-depth analysis of mathematical reasoning abilities of LLMs. All the 17 categories of math word problems are manually annotated with their difficulty levels according to the number of reasoning steps required to solve these problems. We conduct extensive experiments on a wide range of LLMs on FineMath and find that there is still considerable room for improvements in terms of mathematical reasoning capability of Chinese LLMs. We also carry out an in-depth analysis on the evaluation process and methods that have been overlooked previously. These two factors significantly influence the model results and our understanding of their mathematical reasoning capabilities. The dataset will be publicly available soon. 
 \\ \newline  }

\begin{document}

\maketitleabstract

\begin{CJK}{UTF8}{gbsn}
\section{Introduction}
Mathematics has always been an important part of the evaluation of LLMs \citep{DBLP:conf/nips/Wei0SBIXCLZ22}, which not only assesses the ability of LLMs in understanding and solving mathematical problems, but also profoundly measures the essential capability of LLMs in abstract conceptualization, logical reasoning and so on. Therefore, a high-quality mathematical evaluation benchmark is of great importance to a comprehensive LLM evaluation.

Previous works \citep{hosseini-etal-2014-learning,roy-roth-2015-solving} curate mathematical test sets in English, which serve as a repository for grade school math word problems with accuracy being used as the evaluation metric. Recent years have witnessed a substantial progress in Chinese LLMs. Hence, mathematical evaluation datasets in Chinese \citep{DBLP:journals/corr/abs-2306-16636,DBLP:journals/corr/abs-2309-03241} have been created correspondingly. These two previous Chinese datasets categorize testing instances by grade levels, providing a preliminary evaluation of Chinese LLMs on these levels. Their evaluation results show that the accuracy of GPT-4 for any grade surpasses or is close to 60\%. However, a simple accuracy does not help us understand which mathematical concepts or skills LLMs have mastered. There is an urgent need for a comprehensive test set that can provide fine-grained evaluation results.

\begin{figure}[!ht]
\begin{center}
\includegraphics[width=\linewidth]{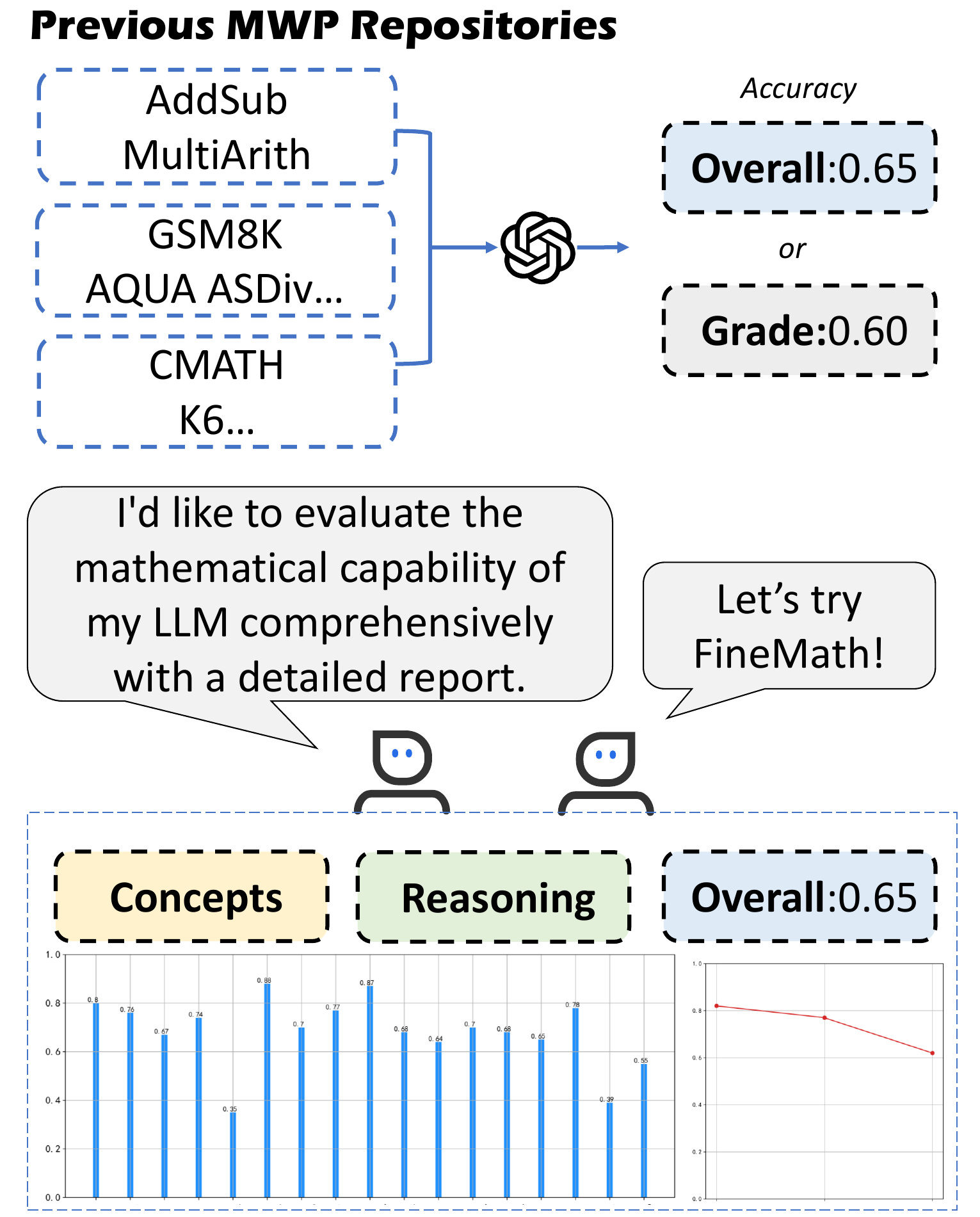} 
\caption{FineMath can evaluate LLMs' mathematical ability from three aspects: accuracy of understanding abstract mathematical concepts, accuracy of reasoning, and overall accuracy.}
\label{fig.1}
\end{center}
\end{figure}

Apart from arithmetic operations, mathematical ability involves diverse reasoning capabilities. We believe that the evaluation of the mathematical ability of LLMs should include two aspects:
\begin{itemize}
    \item Providing diverse abstract mathematical concepts.
    \item Evaluating the logical and mathematic reasoning abilities of LLMs over mathematical problems at different difficulty levels.
\end{itemize}

In pursuit of these aspects, we propose FineMath, a benchmark composed of Math Word Problems (MWPs), designed to comprehensively assess LLMs' mathematical capability in a fine-grained way. FineMath organizes MWPs according to key mathematical concepts taught in elementary school, and each type of MWPs contains three levels of difficulty, facilitating detailed reasoning ability analysis.

Specifically, FineMath consists of 17 types of MWPs. For defining and collecting these MWP types, we have referred to the mathematics curriculum standards established by China's Ministry of Education and the principles and standards for school mathematics set by the American National Council of Teachers of Mathematics (NCTM). The key concepts and skills in grade school include Number \& Operations, Algebra, Geometry, Measurement, Data Analysis \& Probability, Problem Solving, and Reasoning. Different key concepts involve the use of different knowledge and abilities. We have also annotated the reasoning steps and process for each MWP, categorizing them into questions requiring one reasoning step, two reasoning steps, and three or more reasoning steps.

Based on the curated benchmark, we have conducted a thorough analysis of the evaluation process and methods. Evaluations in mathematics have always emphasized the accuracy of results. However, we have observed factors that greatly influence the model's results, thus affecting our understanding of its capabilities:
\begin{itemize}
    \item The model is sensitive to the prompts used during the evaluation, and the results vary accordingly.
    \item The methods of evaluation can also affect the model's results. We have compared the model's performance in selecting the final answers from the options of multi-choice questions, demonstrating that the form of evaluation tasks and options can influence the model's results to a certain extent.
    \item The length of the LLM-generated answers, to some degree, reflects the model's ``confidence'' when handling questions.
\end{itemize}

The main contributions of our work are as follows:
\begin{itemize}
    \item We propose a fine-grained elementary school MWPs benchmark for Chinese LLMs, which can assess the mathematical capabilities of LLMs from three aspects: accuracy of understanding abstract mathematical concepts, accuracy of reasoning, and overall accuracy.
    \item We conduct an in-depth analysis of the contamination in our dataset, enabling researchers of LLMs to conduct a credibility analysis on the evaluation results.
    \item We evaluate GPT-4, GPT-3.5-Turbo, and 8 Chinese LLMs, revealing their mathematical reasoning capabilities, and provide detailed evaluation results in various aspects.
\end{itemize}

\begin{table}
\centering
\resizebox{\linewidth}{!}{
\begin{tabular}{lcc}
\toprule
\textbf{Test Sets} & \textbf{Size} & \textbf{Language}  \\
\midrule
AddSub \citep{hosseini-etal-2014-learning} & 395 & En  \\
MultiArith \citep{roy-roth-2015-solving} & 600 & En  \\
SingleEq \citep{DBLP:journals/tacl/Koncel-Kedziorski15} & 508 & En \\
AQUA \citep{ling-etal-2017-program} & 100K & En  \\
AsDiv \citep{miao-etal-2020-diverse} & 2,305 & En  \\
GSM8K \citep{DBLP:journals/corr/abs-2110-14168} & 8.5k & En  \\
SVAMP \citep{DBLP:conf/naacl/PatelBG21} & 8.5k & En  \\
\midrule
CMATH \citep{DBLP:journals/corr/abs-2306-16636} & 1.7K & Zh  \\
K6 \citep{DBLP:journals/corr/abs-2309-03241} & 600 & Zh  \\
FineMath (ours) & 1,584 & Zh \\
\bottomrule
\end{tabular}}
\caption{\label{related-work} An overview of MWP datasets.}
\end{table}

\section{Related Work}
Traditional MWP datasets like AddSub \citep{hosseini-etal-2014-learning} and MultiArith \citep{roy-roth-2015-solving} are integrated into a MWP repository. Other similar datasets include SingleEq \citep{DBLP:journals/tacl/Koncel-Kedziorski15}, AQUA \citep{ling-etal-2017-program} and AsDiv \citep{miao-etal-2020-diverse}. GSM8K \citep{DBLP:journals/corr/abs-2110-14168} and SVAMP \citep{DBLP:conf/naacl/PatelBG21} take advantage of detailed annotations and have prevailed in recent evaluations.

Our work is most inspired by the MATH \citep{DBLP:conf/nips/HendrycksBKABTS21} dataset. MATH collects problems from American high school mathematics competitions and categorizes problems into seven subjects. These subjects are Prealgebra, Algebra, Number Theory, Counting and Probability, Geometry, Intermediate Algebra and Precalculus. However, these problems are very challenging, even when humans answer them, the accuracy rate is only 40\%.  Considering that many LLMs are still in their early versions, overly difficult problems may have limited significance for testing these models.

CMATH proposed by \citep{DBLP:journals/corr/abs-2306-16636} and K6 proposed by \citep{DBLP:journals/corr/abs-2309-03241} are the two datasets that are relatively similar to ours developed concurrently. All these datasets focus on math word problems of elementary school, and organize instances by grade level. CMATH contains 1.7K problems collected from workbooks and exams on the Internet. K6 is composed of 600 problems collected from an educational institution. However, neither of the two datasets have been publicly released, precluding us from conducting an empirical comparison to them.

Ape210K proposed in \citep{zhao2020ape210k} is a slightly earlier dataset. It contains 210K enormous Chinese math word problems from elementary school. Test sets alone in Ape210K contain as many as 5,000 problems. However, the test sets do not provide annotations related to LLMs.

An overview of the related MWP datasets is shown in Table~\ref{related-work}.

% \begin{table}
% \centering
% \resizebox{\linewidth}{!}{
% \begin{tabular}{lcccc}
% \toprule
% \textbf{Type} & \textbf{Total} & \textbf{Level-1} & \textbf{Level-2} & \textbf{Level-3} \\
% \midrule
% Percents & 60 & 20 & 20 & 20 \\
% Decimals & 93 & 23 & 30 & 40 \\
% Fractions & 81 & 24 & 27 & 30 \\
% Fac\&Multi & 61 & 20 & 21 & 20 \\
% Counting & 60 & 20 & 20 & 20 \\
% Proportions & 78 & 20 & 20 & 38 \\
% Mix Operations & 267 & 91 & 107 & 69 \\
% \midrule
% Spatial Sense & 89 & 20 & 47 & 22 \\
% Time & 64 & 20 & 24 & 20 \\
% \midrule
% Central Tendency & 189 & 22 & 98 & 69 \\
% Probability & 68 & 28 & 20 & 20 \\
% \midrule
% Equations & 100 & 0 & 25 & 75 \\
% Patterns & 60 & 20 & 20 & 20 \\
% \midrule
% Basic Geometry & 132 & 20 & 48 & 64 \\
% Analytic Geometry & 60 & 20 & 20 & 20 \\
% \midrule
% Problem 1 & 62 & 20 & 21 & 21 \\
% Problem 2 & 60 & 20 & 20 & 20 \\
% \bottomrule
% \end{tabular}}
% \caption{\label{detailed-information} Overall statistics of FineMath. Level-1/2/3 denotes that a math word problem requires 1/2/3+ mathematical reasoning steps to solve. }
% \end{table}

\section{Data Collection and Annotation}
We create our dataset by collecting a diverse set of questions. We collect as many questions as possible from textbooks, workbooks and the Internet, from which high-quality questions are selected.

After collecting these questions, we conduct automatical preprocessing on the collected data, which includes removing questions that are not math word problems, discarding questions with fewer than 10 Chinese characters, and retaining only questions that contain definite answers. Additionally, any questions that require reference to images are also discarded.

On the preprocessed data, we further perform manual annotation and processing: MWP categorization, question standardization, reasoning step and answer standardization and multiple-choice question transformation. We elaborate these data curation steps in the following subsections. 

\subsection{MWP Categorization}
We categorize the collected questions into 17 types, each corresponding to a key or basic concept\footnote{https://www.nctm.org/Standards-and-Positions/Principles-and-Standards/Principles,-Standards,-and-Expectations/} inherent in the MWPs. We introduce these key concepts encompassed in our dataset, along with their corresponding categories as follows.

\textbf{Number \& Operations}: This mathematical concept requires an understanding of numbers, ways of representing numbers, relationships among numbers, and number systems. It also necessitates an understanding of the meanings of operations and the ability to compute with these operations. This concept includes 7 MWP categories: Percents, Decimals, Fractions, Factors \& Multiples, Counting, Proportions and Mixed Operations.

\textbf{Measurement}: Measurement requires an understanding of the measurable attributes of objects and the units, systems, and processes of measurement. It corresponds to two MWP categories, namely Spatial Sense and Time.

\textbf{Data Analysis \& Probability}: Data analysis and probability requires one to select and use appropriate statistical methods to analyze data and to apply basic concepts of probability. This concept is related to Central Tendency and Probability.

\textbf{Algebra}: This concept involves understanding patterns, relations and functions. The MWP categories of this concept include Equations and Patterns.

\textbf{Geometry}: Geometry is to analyze the characteristics and properties of two- and three-dimensional geometric shapes, specify locations and describe spatial relationships using coordinate geometry and other representational systems. It contains two MWP categories: Two \& Three Dimensional Geometry (Basic Geometry) and Analytic Geometry.

\textbf{Others}: We also categorized two types of special MWPs. Problem 1: simple optimization problems. Problem 2: tree planting problems that involve the relationship between points and segments.

\subsection{Question Standardization}
Many questions in the selected data contain multiple queries. We normalize these questions so that each question contains only a single query. Ambiguous queries are also rephrased to enable the model to generate a unique answer. 
% Example can be found in Table~\ref{Standardize-example}.

\subsection{Mathematical Reasoning and Answer Standardization}
The process of answering MWPs is manually conducted, and the ground-truth answers are manually double checked by humans. We ask annotators to provide the steps in answering each MWP. Each step should be atomic and indivisible. For calculations that use a fixed solution formula, e.g., computing the area of a circle, we consider them as single-step MWPs.The reasons for annotating the number of required mathematical reasoning steps are two-fold:

1. The number of required reasoning steps can be treated as a proxy to the difficulty level of MWPs. Intuitively, MWPs that require multiple steps to solve are more difficult than those solved in a single step. The progression from one step to the next also represents the reasoning process. Therefore, we categorize the difficulty of MWPs in our dataset into three levels. MWPs that can be solved in a single step are level-1 MWPs; MWPs that require two steps to solve are level-2 MWPs; level-3 MWPs are those that require three or more steps to solve.

2. Presenting the number of reasoning steps facilitates reviewing and analyzing the collected data, thereby ensuring data quality.

% \begin{table*}
% \centering
% \small
% \begin{tabularx}{0.95\textwidth}{>{\raggedright\arraybackslash}X}
% \toprule
% \textbf{Before} \\
% \midrule
% Q: 小明把某数乘3加上20，误看成某数除以3减去20，得数是72，某数是多少？正确的得数是多少？\\
% \specialrule{0em}{1.5pt}{1.5pt}
% Q: Xiao Ming mistakenly takes a number, multiplies it by 3 and adds 20, as dividing the number by 3 and subtracting 20, and gets 72. What is the number? What is the correct result?\\
% \midrule
% \textbf{After} \\
% \midrule
% Q: 小明把某数乘3加上20，误看成某数除以3减去20，得数是72，正确的得数是多少？\\
% \specialrule{0em}{1.5pt}{1.5pt}
% Q: Xiao Ming mistakenly takes a number, multiplies it by 3 and adds 20, as dividing the number by 3 and subtracting 20, and gets 72. What is the correct result?\\
% \bottomrule
% \end{tabularx}
% \caption{\label{Standardize-example} The .}
% \end{table*}

\subsection{Multiple-Choice Question Transformation}
The original MWPs are accompanied with their single ground-truth answers. To facilitate automatic evaluation, we also transform them into multiple-choice question forms by manually providing additional contrastive answer options, similar to the AQUA dataset \citep{ling-etal-2017-program}. 
%In our experiments, we have observed that some newly developed LLMs are not able to follow instructions well, often generating large chunks of tokens unrelated to the correct answers. Even if the answers can be extracted programmatically, it may still introduce errors. That is the reason why we provide multiple-choice options for each MWP. Apart from the correct answer option, each provided option is designed by the annotators to be as misleading as possible. 
% Examples can be found in Table~\ref{dataset-example}.
%We also provide randomly generated options for comparison, to verify the effectiveness of our transformation. %Examples can be found in Table~\ref{dataset-example}.

\section{Data Statistics and Analysis}
We provide data statistics and analysis on contamination of our dataset in this section.

\subsection{Data Statistics}
\begin{table}
\centering
\resizebox{\linewidth}{!}{
\begin{tabular}{llcccc}
\toprule
\textbf{Concepts} & \textbf{Type} & \textbf{Total} & \textbf{Level-1} & \textbf{Level-2} & \textbf{Level-3} \\
\midrule
\multirow{7}{*}{Number \& Operations} &Percents & 60 & 20 & 20 & 20 \\
&Decimals & 93 & 23 & 30 & 40 \\
&Fractions & 81 & 24 & 27 & 30 \\
&Fac\&Multi & 61 & 20 & 21 & 20 \\
&Counting & 60 & 20 & 20 & 20 \\
&Proportions & 78 & 20 & 20 & 38 \\
&Mix Operations & 267 & 91 & 107 & 69 \\
\midrule
\multirow{2}{*}{Measurement} & Spatial Sense & 89 & 20 & 47 & 22 \\
&Time & 64 & 20 & 24 & 20 \\
\midrule
\multirow{2}{*}{Data Analysis \& Probability} & Central Tendency & 189 & 22 & 98 & 69 \\
& Probability & 68 & 28 & 20 & 20 \\
\midrule
\multirow{2}{*}{Algebra} &Equations & 100 & 0 & 25 & 75 \\
&Patterns & 60 & 20 & 20 & 20 \\
\midrule
\multirow{2}{*}{Geometry} & Basic Geometry & 132 & 20 & 48 & 64 \\
& Analytic Geometry & 60 & 20 & 20 & 20 \\
\midrule
\multirow{2}{*}{Others} &Problem 1 & 62 & 20 & 21 & 21 \\
& Problem 2 & 60 & 20 & 20 & 20 \\
\bottomrule
\end{tabular}}
\caption{\label{detailed-information} Overall statistics of FineMath. Level-1/2/3 denotes that a math word problem requires 1/2/3+ mathematical reasoning steps to solve. }
\end{table}

The overall data statistics are displayed in Table \ref{detailed-information}. All 1,584 questions are categorized into five major mathematical concepts and two classic types of MWPs. Each type contains at least 60 questions, and each difficulty level contains at least 20 questions.

\subsection{Analysis on Contamination}

\begin{figure*}[!ht]
\begin{center}
\includegraphics[width=\textwidth]{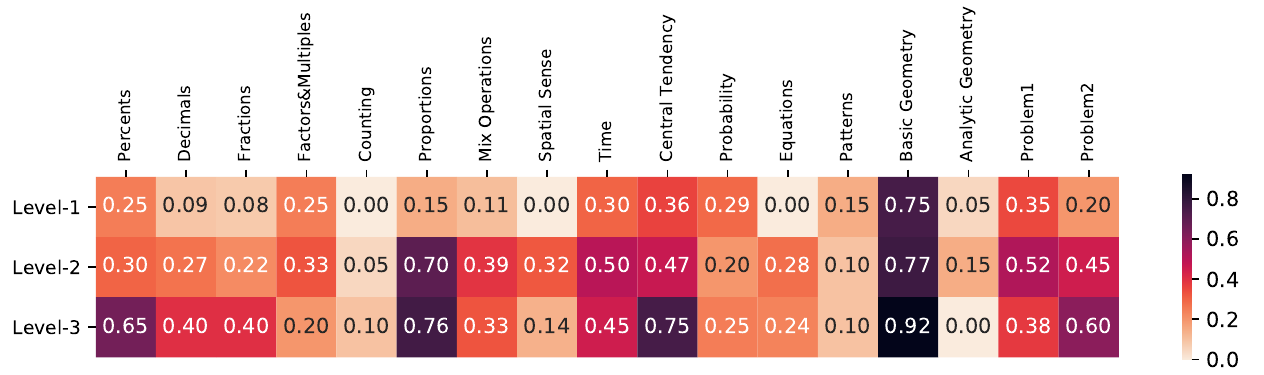} 
\caption{Contamination analysis. The overlap rate between FineMath and the training sets of Ape210k.}
\label{fig.2}
\end{center}
\end{figure*}

FineMath serves as a comprehensive benchmark, encompassing a diverse range of math word problems at the Chinese elementary school level. It is specifically designed to assess the mathematical reasoning capabilities of Chinese large language models. However, these language models are typically trained on a huge amount of data derived from multiple sources, including web pages, books, codes and so on. This raises the potential risk of test data contamination, as some test examples from FineMath may unintentionally be included in the training data of these language models. Test data contamination can lead to an overestimation of a model's performance, potentially resulting in misleading conclusions regarding the model's generalization capabilities. Consequently, investigating contamination and its impact on model performance for FineMath is of paramount importance.

Ape210K \citep{zhao2020ape210k} is a publicly available large-scale Chinese math word problem dataset,  which has been splited into training, validation, and test sets. It is commonly utilized as a training dataset for mathematical problem-solving models \citep{DBLP:journals/corr/abs-2208-05645,DBLP:conf/acl/WuZWH20,DBLP:journals/corr/abs-2307-07951,DBLP:conf/emnlp/HuangWXCY21,DBLP:journals/corr/abs-2210-15373,DBLP:conf/ijcnn/HuangZBROX23,DBLP:journals/corr/abs-2309-03241,DBLP:conf/naacl/LiangZWQLS022}. To determine potential contamination from Ape210K in FineMath, we adopt the identical methodology leveraged in GPT-3 \citep{DBLP:conf/nips/BrownMRSKDNSSAA20} to compute the n-gram overlap between Ape210K and FineMath. In this approach, a test example in FineMath is considered as an overlapped example with Ape210k if any n-gram from this test example also appears in Ape210k. Specifically, we insert white spaces around any Chinese, Japanese, and Korean (CJK) characters, as well as between punctuation marks and words. Subsequently, we tokenize the text based on these white spaces. It is important to note that we disregard letter case when computing n-grams.

To perform a rigorous quantitative assessment of contamination, we define the overlap rate as the fraction of instances within FineMath that exhibit such overlap. Furthermore, for the purposes of computing overlap, we set the value of n to 13. The overlap rate between FineMath and the training sets of Ape210k is depicted in Figure~\ref{fig.2}. It suggests that the overlap rates of some question types are significantly higher than others, such as Basic Geometry and Proportions. To gain deeper insights into the impact of these overlapped examples on model performance, we partition the test examples into two datasets: a contaminated dataset composed of the overlapped examples, and a clean dataset whose test examples exhibit no overlap with the Ape210k training set. Subsequently, we examine the performance of the model on each of these datasets separately. We select GPT-4 and MathGLM-10B for analysis since GPT-4 is widely recognized as the most advanced LLM currently available and MathGLM-10B has been trained on the Ape210k training set which overlaps with some test examples of FineMath. The experimental results are presentend in Table~\ref{contaminate}. Notably, MathGLM-10B performs significantly better on the contaminated dataset compared to the clean dataset. In contrast, GPT-4 exhibits comparable performance on both datasets. This suggests that MathGLM-10B may be overfitting to the overlapped examples and that contamination can inflate a model's performance. Consequently, to ensure a fair comparison between models and to draw accurate conclusions from the FineMath benchmark, we recommend filtering out overlapped examples between the training set and the FineMath benchmark.

% To test whether the model's accuracy on FineMath is affected by contaminated data, we divide the data into two sets: a dirty dataset and a clean dataset, based on the contamination status of each data piece. We then tested the performance of GPT-4 and MathGLM-10B, the latter of which was trained using Ape210K. The details are provided in Table~\ref{contaminate}.

% The results of the 13-gram contamination analysis can be found in the Figure~\ref{fig.2}. The contamination levels for Basic Geometry, Proportions, and Central Tendency are relatively high. To test whether the model's accuracy on FineMath is affected by contaminated data, we divided the data into two sets: a dirty dataset and a clean dataset, based on the contamination status of each data piece. We then tested the performance of GPT-4 and MathGLM-10B, the latter of which was trained using Ape210K. The details are provided in Table~\ref{contaminate}.

% The test results reveal that MathGLM-10B performs better on problem types with severe contamination, and there is a significant difference in its scores between the clean and contaminated datasets. In contrast, GPT-4's performance does not appear to be affected by contaminated data, as it scores consistently on both datasets.

\begin{table}
\centering
\resizebox{\linewidth}{!}{
\begin{tabular}{lcccc}
\toprule
\textbf{Model/Dataset} & \textbf{Level-1} & \textbf{Level-2} & \textbf{Level-3} & \textbf{Overall} \\
\midrule
\textbf{MathGLM-10B (Clean)} & 0.45 & 0.42 & 0.22 & 0.37 \\
\textbf{MathGLM-10B (Contaminated)} & 0.65 & 0.75 & 0.70 & 0.71 \\
\midrule
\textbf{GPT-4 (Clean)} & 0.83 & 0.76 & 0.61 & 0.74 \\
\textbf{GPT-4 (Contaminated)} & 0.83 & 0.65 & 0.63 & 0.67 \\
\bottomrule
\end{tabular}}
\caption{\label{contaminate} Accuracy results of GPT-4 and MathGLM-10B on the contaminated dataset and clean dataset.}
\end{table}

\section{Experiments}
We conducted experiments on the proposed FineMath to evaluate a series of LLMs, assessing the mathematical reasoning capabilities of them. 

\subsection{Evaluated LLMs}
We assessed three classes of LLMs: GPT-4 and GPT-3.5-Turbo developed by OpenAI; LLMs developed for Chinese; and LLMs finetuned with Chinese mathematics data. Specific information can be found in Table~\ref{models}.

\begin{table}
\centering
\resizebox{\linewidth}{!}{
\begin{tabular}{lcccc}
\toprule
\textbf{Model} & \textbf{RLHF} & \textbf{Parameters} & \textbf{Training Token} \\
\midrule
\textbf{GPT-4} & w & - & -  \\
\textbf{GPT-3.5-Turbo} & w & - & -  \\
\textbf{GhatGLM2-6B} & w & 6B & 1.4T  \\
\textbf{Moss-SFT-16B} & w/o & 16B & 120B  \\
\textbf{InternLM-Chat-7B} & w & 7B & 1.6T  \\
\textbf{Qwen-7B-Chat} & w & 7B & 2.4T  \\
\textbf{Baichuan-7B} & w/o & 7B & 1.2T  \\
\textbf{Baichuan2-7B-Chat} & w & 7B & 2.6T  \\
\textbf{MathGLM-10B} & - & 10B & -  \\
\textbf{MathGLM-335M} & - & 335M & -  \\
% \textbf{Chinese-Alpaca2-7B} & w & 7B & - & LLaMA &  \\
% \textbf{Chinese-Alpaca2-13B} & w & 13B & - & LLaMA &  \\
\bottomrule
\end{tabular}}
\caption{\label{models} All the LLMs that we evaluated in this paper. MathGLM-10B and MathGLM-335M, both of which were fine-tuned using an arithmetic training dataset.}
\end{table}

\begin{table}
\centering
\resizebox{\linewidth}{!}{
\begin{tabular}{ll}
\toprule
\multirow{2}{*}{\textbf{Prompt 0:} }& \multirow{2}{*}{Nothing is provided, only the question is input into the model.} \\
                                    & \\
                                    \midrule
\multirow{2}{*}{\textbf{Prompt 1:}} & Here is a math word problem; please provide the answer to this question. \\ 
                                    & Do not explain the reason. \\
                                    \midrule
\multirow{2}{*}{\textbf{Prompt 2:}} & Here is a math word problem; please select the correct option. \\                                                & Do not explain the reason. \\
\midrule
\multirow{2}{*}{\textbf{Prompt 3:}} & Here is a math word problem, please give the answer to this question. \\
                                    & And explain why. \\
                                    \midrule
\multirow{2}{*}{\textbf{Prompt 4:}} & \multirow{2}{*}{Answer:....} \\
& \\
\bottomrule
\end{tabular}}
\caption{\label{Prompts} Prompts used for evaluation and analysis.}
\end{table}

\subsection{Prompts}
All experiments were conducted under the zero-shot. We tried several prompts for evaluation and analysis, which are shown in Table~\ref{Prompts}.

\subsection{Main Results}

\begin{figure}[t]
\begin{center}
\includegraphics[width=\linewidth]{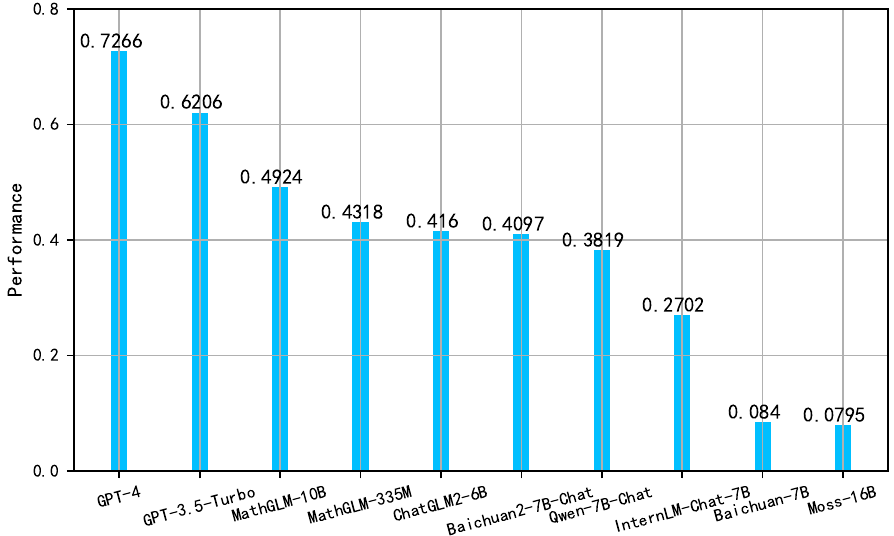} 
\caption{Main results of different evaluated LLMs on our dataset (under Prompt 0).}
\label{fig.3}
\end{center}
\end{figure}

\begin{table*}
\centering
\resizebox{\textwidth}{!}{
\begin{tabular}{lccccccccccccccccc}
\toprule
\multirow{2}{*}{\textbf{Model}} & \multirow{2}{*}{\textbf{Percents}} & \multirow{2}{*}{\textbf{Decimals}} & \multirow{2}{*}{\textbf{Fractions}} & \textbf{Factors \&} & \multirow{2}{*}{\textbf{Counting}} & \multirow{2}{*}{\textbf{Proportions}} & \textbf{Mix} & \textbf{Spatial} & \multirow{2}{*}{\textbf{Time}} & \textbf{Central} & \multirow{2}{*}{\textbf{Probability}} & \multirow{2}{*}{\textbf{Equations}} & \multirow{2}{*}{\textbf{Patterns}} & \textbf{Basic} & \textbf{Analytic} & \multirow{2}{*}{\textbf{Problem 1}} & \multirow{2}{*}{\textbf{Problem 2}} \\
& & & & \textbf{Multiples} & & & \textbf{Operations} & \textbf{Sense} &  & \textbf{Tendency} & & & &\textbf{Geometry} & \textbf{Geometry} & & \\
\midrule
GPT-4 & \textbf{0.8} & \textbf{0.76} & 0.67 & \textbf{0.74} & \textbf{0.38} & \textbf{0.78} & \textbf{0.89} & \textbf{0.71} & \textbf{0.77} & \textbf{0.87} & \textbf{0.68} & 0.64 & \textbf{0.68} & 0.7 & \textbf{0.68} & 0.39 & 0.55 \\
GPT-3.5-Turbo & 0.75 & 0.7 & \textbf{0.69} & 0.7 & 0.33 & 0.65 & 0.81 & 0.57 & 0.69 & 0.71 & 0.34 & \textbf{0.71} & 0.47 & 0.49 & 0.42 & 0.23 & 0.55 \\
ChatGLM2-6B & 0.65 & 0.46 & 0.3 & 0.52 & 0.3 & 0.33 & 0.63 & 0.63 & 0.38 & 0.35 & 0.24 & 0.38 & 0.2 & 0.46 & 0.18 & 0.1 & 0.3 \\
Moss-SFT-16B & 0.13 & 0.05 & 0.07 & 0.11 & 0.03 & 0.06 & 0.15 & 0.06 & 0.09 & 0.03 & 0.12 & 0.05 & 0.07 & 0.05 & 0.03 & 0.08 & 0.05\\
InternLM-Chat-7B & 0.4 & 0.35 & 0.15 & 0.34 & 0.05 & 0.31 & 0.42 & 0.36 & 0.23 & 0.17 & 0.18 & 0.23 & 0.22 & 0.18 & 0.25 & 0.21 & 0.32 \\
Qwen-7B-Chat & 0.58 & 0.53 & 0.41 & 0.39 & 0.1 & 0.32 & 0.58 & 0.42 & 0.28 & 0.31 & 0.35 & 0.34 & 0.32 & 0.38 & 0.2 & 0.18 & 0.25 \\
Baichuan-7B & 0.08 & 0.13 & 0.01 & 0.08 & 0.05 & 0.03 & 0.14 & 0.08 & 0.05 & 0.1 & 0.12 & 0.06 & 0.07 & 0.07 & 0.07 & 0.05 & 0.08\\
Baichuan2-7B-Chat & 0.58 & 0.52 & 0.48 & 0.54 & 0.2 & 0.4 & 0.65 & 0.58 & 0.25 & 0.32 & 0.26 & 0.4 & 0.25 & 0.37 & 0.13 & 0.11 & 0.22\\
MathGLM-10B & 0.52 & 0.42 & 0.68 & 0.57 & 0.05 & 0.64 & 0.61 & 0.38 & 0.45 & 0.53 & 0.29 & 0.36 & 0.18 & \textbf{0.74} & 0.23 & \textbf{0.4} & \textbf{0.65}\\
MathGLM-335M & 0.45 & 0.3 & 0.63 & 0.49 & 0.07 & 0.47 & 0.57 & 0.34 & 0.3 & 0.47 & 0.29 & 0.31 & 0.08 & 0.68 & 0.23 & 0.39 & 0.58 \\
% Chinese-Alpaca2-7B & & & & & & & & & & & & & & & & & \\
% Chinese-Alpaca2-13B & & & & & & & & & & & & & & & & & \\
\bottomrule
\end{tabular}}
\caption{\label{17-type} Results across the 17 MWP categories (under Prompt 0).}
\end{table*}

\begin{figure}[t]
\begin{center}
\includegraphics[width=\linewidth]{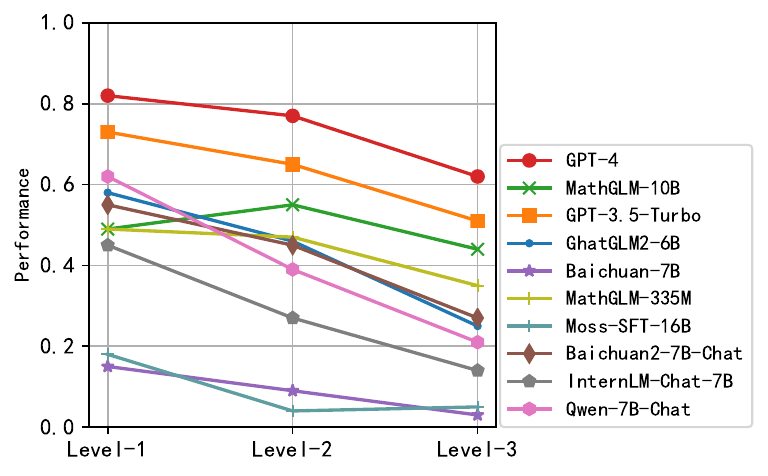} 
\caption{Results in terms of the number of mathematical reasoning steps (under Prompt 0).}
\label{fig.5}
\end{center}
\end{figure}

The overall accuracy results of assessed LLMs are visualized in Figure~\ref{fig.3}. GPT-4 and GPT-3.5-turbo perform outstandingly, with their accuracies reaching as high as 73\% and 62\%, respectively.

Among the evaluated Chinese LLMs, MathGLM-10B, MathGLM-335M, ChatGLM2-6B and Baichuan2-7B-Chat obtain an accuracy of > 40\%. Qwen-7B-Chat and InternLM-Chat-7B are at a slightly below-average level. However, both Baichuan-7B and Moss-SFT-16B perform poorly on our dataset, with an accuracy of < 10\%. Upon examining the responses generated by these two models, we find that their answers often stray from the MWPs, generating a lot of irrelevant content or repeatedly producing the same questions.

By considering both model accuracy and the detailed information provided in Table~\ref{models}, we deduce that the lower accuracy of Moss-SFT-16B is due to an insufficient amount of training data. The performance of Baichuan-7B is hampered because it has not undergone RLHF fine-tuning, which prevents the model from fully understanding the question. In contrast, the accuracy of Baichuan2-7B-Chat, which has been fine-tuned, has significantly improved. In summary, RLHF fine-tuning, having model parameters exceeding 6 billion, and training data reaching the trillion level are all crucial for training an LLM with problem-solving and reasoning capabilities.

% Notably, MathGLM-335M, which was specifically trained for mathematical ability, demonstrated a performance that can compete with ChatGLM2-6B, despite having only 335 million parameters compared to ChatGLM2-6B's 6 billion.

\subsection{Results across 17 MWP Categories}
Results across the 17 MWP categories are displayed in Table \ref{17-type}. It is evident that the MWP types ``Counting'' and ``Problem1'' are more challenging than other MWP categories according to the results. This could be due to the complexities involved in the counting of Chinese numerals and their conversion to Arabic numerals, and the common sense issues encountered in the optimization problem for ``Problem1''. All models demonstrate better performance on ``Mixed Operation'' than on other categories. We also observe that the performance of different models vary significantly.

GPT-4 outperforms all other models. It achieves an accuracy below 40\% on only two MWP categories, surpasses 80\% on three MWP categories, exceeds 70\% in six MWP categories, and goes beyond 60\% in five MWP categories. GPT-4 scores the highest, 89\%, on the 'Mixed Operation' MWPs, while it scores the lowest, 38\%, on the 'Counting' MWPs. Notably, GPT-4 excels in Probability and Analytic Geometry, outperforming all other models with an improvement of over 25\%. GPT-3.5-Turbo's performance on different MWP categories is similar to that of GPT-4, but there is a significant difference (over 20\%) on Probability, Basic Geometry, and Analytic Geometry.

MathGLM-10B and ChatGLM2-6B exhibit similar performance trends across most MWP categories. However, significant differences are observed on Fractions, Counting, Spatial Sense, Basic Geometry, Proportions, Problem1, and Problem2. Due to contamination, MathGLM-10B surpasses GPT-4 on Basic Geometry, Problem 1 and Problem 2.

Baichuan2-7B-Chat, Qwen-7B-Chat and InternLM-Chat-7B are still at a medium or slightly below-average level. Both Moss-SFT-16B and Baichuan-7B perform poorly across all MWP categories, failing to achieve an accuracy of 20\%.

\begin{table}
\centering
\resizebox{\linewidth}{!}{
\begin{tabular}{lcccc}
\toprule
\textbf{Prompt} & \textbf{Level-1} & \textbf{Level-2} & \textbf{Level-3} & \textbf{Overall} \\
\midrule
\textbf{Prompt 0} & 0.83 & 0.77 & 0.62 & 0.73 \\
\midrule
\textbf{Prompt 1} & 0.81 & 0.69 & 0.34 & 0.59 \\
\midrule
\textbf{Prompt 4} & 0.8 & 0.65 & 0.36 & 0.58 \\
\bottomrule
\end{tabular}}
\caption{\label{gpt4-response} Accuracy results of GPT-4 with Prompt 0, Prompt 1, Prompt 4.}
\end{table}

\begin{table*}
\centering
\small
\begin{tabularx}{\textwidth}{llXX}
\toprule
\textbf{Question} &
&学校买来120本课外读物，其中的40\%分给高年级，剩下的课外读物按5:3分给中年级和低年级，低年级分到多少本?&The school bought 120 supplementary books, 40\% of which were given to the upper grades. The remaining books were divided between the middle and early grades in a ratio of 5:3. How many books did the early grades receive?  \\
\midrule
\textbf{Prompt 0+Question} & &\textbf{GPT-4 Response:} 高年级分到的课外读物为120*40\%=48本;剩下的课外读物为120-48=72本;低年级分到的课外读物为72*3/(5+3)=27本。答：低年级分到27本。 \textbf{Results:} \usym{2714} & \textbf{GPT-4 Response:} The number of supplementary books allocated to the upper grades is 120*40\%=48 books. The remaining books total 120-48=72 books. The supplementary books allocated to the early grades are calculated as 72*3/(5+3)=27 books. Answer: The early grades are allocated 27 books.  \\
\midrule
\textbf{Prompt 1+Question} & &\textbf{GPT-4 Response:} 36本 \textbf{Results:} \usym{2717} & \textbf{GPT-4 Response:} 36 books  \\
\midrule
\textbf{Prompt 3+Question} & &\textbf{GPT-4 Response:} 36本。首先，学校将120本书的40\%分给高年级，即120*40\%=48本。剩下的书就是120-48=72本。然后，这72本书按照5:3的比例分给中年级和低年级。这个比例的总和是5+3=8，所以低年级分到的书就是72*(3/8)=27本。\textbf{Results:} \usym{2717} & \textbf{GPT-4 Response:} 36 books. Firstly, the school allocates 40\% of the 120 books to the upper grades. This totals to 120*40\%=48 books. The remaining books amount to 120-48 = 72 books. Then, these 72 books are divided between the middle and early grades at a ratio of 5:3. The total of this ratio is 5+3=8. Therefore, the number of books allocated to the early grades is 72*(3/8) =27 books.  \\
\midrule
\textbf{Prompt 4+Question} & &\textbf{GPT-4 Response:} 36本 \textbf{Results:} \usym{2717} & \textbf{GPT-4 Response:} 36 books  \\
\midrule
\textbf{Prompt 0+Question} & &\textbf{GPT-4 Response:} C:27 & \textbf{GPT-4 Response:} C:27  \\
 (with MCQs) &  &  & \\
 A:72 B:48 C:27 D:80 & & \textbf{Results:} \usym{2714} & \\
 \midrule
\textbf{Prompt 0+Question} & &\textbf{GPT-4 Response:} D:36 & \textbf{GPT-4 Response:} D:36  \\
 (with \textbf{D} replaced) &  &  & \\
 A:72 B:48 C:27 \textbf{D:36} & & \textbf{Results:} \usym{2717} & \\
\bottomrule
\end{tabularx}
\caption{\label{prompts-results}Different prompts and their corresponding responses from GPT-4. MCQs: multiple-choice questions. }
\end{table*}

\subsection{Results in terms of the Number of Mathematical Reasoning Steps}
We further conducted experiments to evaluate LLMs reasoning ability in terms of the number of mathematical reasoning steps. Results are shown in Figure~\ref{fig.5}. We observe that the performance of LLMs decreases as the number of reasoning steps increases. GPT-4 maintains an accuracy of over 60\% at all difficulty levels, reaching as high as 82\% on MWPs that require only one step of reasoning. The accuracy of GPT-3.5-Turbo is, on average, 10\% lower than that of GPT-4. While ChatGLM2-6B, Baichuan2-7B-Chat and Qwen-7B-Chat outperform MathGLM-335M and MathGLM-10B on Level-1 MWPs, its accuracy falls below those of MathGLM-335M and MathGLM-10B on Level-2/3 MWPs. Similar to their performance across MWP categories, Moss-SFT-16B and Baichuan-7B show a significant difference in performance at all difficulty levels compared to the other models.

The accuracy difference between Qwen-7B-Chat and InternLM-Chat-7B on different reasoning steps is quite substantial, exceeding 30\%. In the case of Qwen-7B-Chat, the accuracy on problems requiring only one-step reasoning is 62\%, but this figure drops to just 21\% for problems requiring three or more reasoning steps. This phenomenon suggests that the model may need more training in terms of inference.

\section{Analysis}
Unlike other studies that only evaluate accuracy, we have further analyzed factors in the evaluation process. These overlooked factors greatly affect the evaluation results and our understanding of the true mathematical reasoning capabilities of LLMs.

\subsection{Prompts Really Does Matter}
During the evaluation, instructions are generally used to guide the assessed model to produce answers. For instance, we might say, ``Here is a math problem, please provide the answer to this question. Do not explain the reason.'' Alternatively, we might provide an answer template such as ``Question: ... Answer:''. However, our experiments showed that even a single word like ``Answer:'' can significantly affect the model's accuracy. We tested three prompt 0, prompt 1 and Prompt4 on GPT-4, results are shown in Table~\ref{gpt4-response}.

We can see that the overall accuracy results with the three prompts are 73\%, 59\%, and 58\%, respectively, with a gap reaching up to 15\%.

Prompt like ``Answer:'' appear to encourage the model to forego reasoning and directly provide the answer, which increases the likelihood of generating incorrect responses. An example is shown in Table~\ref{prompts-results}: Prompt 4+Question.

\subsection{Evaluation Methods: Generation vs. Option Prediction}
In our preliminary experiments, we have discovered that some newly developed LLMs do not follow instructions well, often generating large chunks of tokens unrelated to the answer. Therefore, we decide to transform our data into multiple-choice questions, for which the evaluated model can then select the correct answer option. 
%We conducted transformation in two ways: manually creating answer options in addition to the correct answer and %randomly generating answer options.

% \begin{table}
% \centering
% \resizebox{\linewidth}{!}{
% \begin{tabular}{lcccc}
% \toprule
% \textbf{Evaluation Method} & \textbf{Level-1} & \textbf{Level-2} & \textbf{Level-3} & \textbf{Overall} \\
% \midrule
% \textbf{Prompt 0} & 0.82 & 0.77 & 0.62 & 0.73 \\
% \midrule
% \textbf{Prompt 0 (with MCQs)} & 0.67 & 0.6 & 0.6 & 0.62 \\
% \bottomrule
% \end{tabular}}
% \caption{\label{gen-choice} Accuracy of GPT-4 with different evaluation methods: generation vs. option prediction.}
% \end{table}

\begin{table}
\centering
\resizebox{\linewidth}{!}{
\begin{tabular}{lccccc}
\toprule
\textbf{Model} & \textbf{MCQs} & \textbf{Level-1} & \textbf{Level-2} & \textbf{Level-3} & \textbf{Overall} \\
\midrule
\multirow{2}{*}{GPT-4} & w/o & 0.83 & 0.77 & 0.62 & 0.73 \\
& w & 0.67 & 0.6 & 0.6 & 0.62 \\
\midrule
\multirow{2}{*}{GPT-3.5-Turbo} & w/o & 0.73 & 0.65 & 0.51 & 0.62 \\
& w & 0.65 & 0.66 & 0.57 & 0.63 \\
\midrule
\multirow{2}{*}{ChatGLM2-6B} & w/o & 0.58 & 0.46 & 0.25 & 0.42 \\
& w & 0.69 & 0.62 & 0.46 & 0.58 \\
\midrule
\multirow{2}{*}{Moss-SFT-16B} & w/o & 0.18 & 0.04 & 0.05 & 0.08 \\
& w & 0.16 & 0.13 & 0.14 & 0.14 \\
\midrule
\multirow{2}{*}{InternLM-Chat-7B} & w/o & 0.45 & 0.27 & 0.14 & 0.27 \\
& w & 0.29 & 0.26 & 0.29 & 0.28 \\
\midrule
\multirow{2}{*}{Qwen-7B-Chat} & w/o & 0.62 & 0.39 & 0.21 & 0.38 \\
& w & 0.47 & 0.42 & 0.39 & 0.42 \\
\midrule
\multirow{2}{*}{Baichuan-7B} & w/o & 0.15 & 0.09 & 0.03 & 0.08 \\
& w & 0.32 & 0.25 & 0.25 & 0.27 \\
\midrule
\multirow{2}{*}{Baichuan2-7B-Chat} & w/o & 0.55 & 0.45 & 0.27 & 0.41 \\
& w & 0.44 & 0.42 & 0.32 & 0.39 \\
% \midrule
% \multirow{2}{*}{Moss-SFT-16B} & w/o &  &  &  &  \\
% & w &  &  &  &  \\
\bottomrule
\end{tabular}}
\caption{\label{gen-choice} Accuracy of LLMs with different evaluation methods: generation vs. option prediction.}
\end{table}

Comparison results are displayed in Table~\ref{gen-choice}. We can observe a significant difference in accuracy between option prediction (with multiple-choice questions) and direct answer generation, with a gap that can exceed 10\%. Interestingly, restructuring the task in the form of multiple-choice questions seems to reduce the accuracy of high-performing models while increasing the accuracy of models that perform poorly. Upon examining instances, we have found that the answer option can act as another type of prompt influencing the model's performance. For example in Table~\ref{prompts-results}: Prompt 0+Question (with MCQs) and Prompt 0+Question (with \textbf{D} replaced).

Examples of GPT-4 outputs with different prompts and task forms are shown in Table~\ref{prompts-results}. We want to understand why GPT-4 would provide the incorrect answer ``36'' under Prompt 1. Therefore, we utilize Prompt 3 to have GPT-4 explain its reasoning for choosing ``36''. Interestingly, GPT-4 mentions the correct number ``27'' in the explanation, but still provides the incorrect answer, ``36''. Given the seeming importance of ``36'', we replace one of the options in the multiple-choice question with ``36''. The model, which initially selected the correct answer, abandoned it in favor of ``36''. Further examples can be observed in the responses generated by other models. For instance, the reasoning process may be incorrect, yet the correct answer is ultimately chosen. The options provided to the model also seem to influence the model's generation probability to a certain degree. Therefore, we recommend using generation, rather than option prediction, for a more accurate evaluation of LLMs.

\subsection{Comparison of Response Lengths}
We conducted a statistical analysis on the length of the responses generated by the models. We discover two phenomena. First, models like GPT-4 and GPT-3.5-Turbo tend to generate responses that are closely centered around the question, with shorter text. This may demonstrate the characteristics of models with high accuracy. Second, the more reasoning steps an MWP requires, the longer the response tends to be. Please refer to Table~\ref{response-length} for more details.

\begin{table}
\centering
\resizebox{\linewidth}{!}{
\begin{tabular}{lccc}
\toprule
\textbf{Model} & \textbf{Level-1} & \textbf{Level-2} & \textbf{Level-3} \\
\midrule
GPT-4 & 33.18 & 56.92 & 102.37 \\
GPT-4+Prompt 4 & 8.67 & 13.12 & 26.73 \\
GPT-3.5-Turbo & 71.10 & 104.78 & 173.87 \\
GhatGLM2-6B & 167.75 & 224.02 & 381.18 \\
InternLM-Chat-7B & 65.65 & 72.20 & 119.69 \\
Qwen-7B-Chat & 62.29 & 93.57 & 138.21 \\
Baichuan2-7B-Chat & 114.72 & 156.33 & 202.52 \\
Baichuan-7B & 128.48 & 174.73 & 130.21 \\
Moss-SFT-16B & 105.43 & 131.29 & 149.48 \\
\bottomrule
\end{tabular}}
\caption{\label{response-length} Different response lengths of models (Since MathGLM-10B and MathGLM-335M, after fine-tuning, generates more formulas but fewer language descriptions,  is not compared in this context).}
\end{table}

We speculate that the model's ``confidence'' in answering questions influences the length of its response. This tendency is also observed in some models that do not adhere strictly to instructions. For instance, even when instructed to provide only the answer without explaining, these models still generate logical reasoning for difficult MWPs.

\section{Conclusion}
We present a fine-grained benchmark, FineMath, to comprehensively evaluate the mathematical capabilities of Chinese LLMs. We strive to evaluate as many LLMs as possible. We also conduct a contamination analysis, enabling researchers to examine whether the training data influences the evaluation results. Through testing various evaluation methods and processes, we demonstrate their potential to influence the results, underscoring the necessity for fair and effective evaluations to consider the interference caused by these two aspects.

\end{CJK}

\nocite{*}
\section{Bibliographical References}\label{sec:reference}

\bibliographystyle{lrec-coling2024-natbib}
\bibliography{lrec-coling2024-example}

% \section{Language Resource References}
% \label{lr:ref}
% \bibliographystylelanguageresource{lrec-coling2024-natbib}
% \bibliographylanguageresource{languageresource}

\end{document}